\documentclass[journal,transmag]{IEEEtran}
\usepackage{textcomp} 
\usepackage{amsmath}
\usepackage{pgfplots}
\usetikzlibrary{patterns}
\usepackage{siunitx}

\usepackage{cite}
\begin{document}
\title{Real-time Local Noise Filter in 3D Visualization of CT Data}


\author{\IEEEauthorblockN{Nicholas Tan Jerome\IEEEauthorrefmark{1},
Zhassulan Ateyev\IEEEauthorrefmark{2},
Sebastian Schmelzle\IEEEauthorrefmark{3}, 
Suren Chilingaryan\IEEEauthorrefmark{1}, and
Andreas Kopmann\IEEEauthorrefmark{1}}
\IEEEauthorblockA{\IEEEauthorrefmark{1}Institute for Data Processing and Electronics,
Karlsruhe Institute of Technology, Germany}
\IEEEauthorblockA{\IEEEauthorrefmark{2}School of Computer Science \& Robotics, Tomsk Polytechnic University, Russia}
\IEEEauthorblockA{\IEEEauthorrefmark{3}Ecological Networks, Department of Biology, Technische Universit\"{a}t Darmstadt, Germany}

%\thanks{Manuscript received June XX, 2018; revised XXXX, 2018. 
%Corresponding author: N. Tan Jerome (email: nicholastanjerome@gmail.com)}

}

%{Shell \MakeLowercase{\textit{et al.}}: Bare Demo of IEEEtran.cls for IEEE Transactions on Magnetics Journals}



% for Transactions on Magnetics papers, we must declare the abstract and
% index terms PRIOR to the title within the \IEEEtitleabstractindextext
% IEEEtran command as these need to go into the title area created by
\IEEEtitleabstractindextext{%
\begin{abstract}
Removing noise in computer tomography (CT) data for real-time 3D visualization is vital to improving the quality of the final display. However, the CT noise cannot be removed by straight averaging because the noise has a broadband spatial frequency that is overlapping with the interesting signal frequencies. To improve the display of structures and features contained in the data, we present spatially variant filtering that performs averaging of sub-regions around a central region. We compare our filter with four other similar spatially variant filters regarding entropy and processing time. The results demonstrate significant improvement of the visual quality with processing time still within the millisecond range.
\end{abstract}

}

\maketitle

\IEEEdisplaynontitleabstractindextext

% \fi
\IEEEpeerreviewmaketitle

\section{Introduction}

\IEEEPARstart{I}{n} recent years, X-ray computed tomography (CT) imaging allows biologists to study the internal structure of small animals such as insects and other arthropods~\cite{betz2007imaging,schmelzle2015mechanics,van2013insect}. However, noise is inevitable in data produced by CT. It arises from various sources such as photon detection statistics, detector misalignment, reconstruction algorithms, and so on~\cite{borsdorf2009adaptive,boas2012ct}. Hence, the visualization of the original CT data with the inherited noise obscures the user from identifying the structures and features contained in the data. Even when the user filters the information manually, essential details could be lost due to the broad spectral range of the CT noise, overlapping with the signal frequencies of interest~\cite{okada1985noise}.

% concomitant noise

Most research emphasizes offline data processing performing data segmentation for a better understanding of the inner data structure. L\"{o}sel and Heuveline~\cite{losel2016enhancing} described a semi-automated segmentation approach based on the random walk algorithm. As a result, noise is removed by the time-consuming segmentation process. In this paper, we aim to provide a fast preview of the original CT data. We focus on real-time rendering where the noisy data must be processed within the millisecond range. Note that we do not remove any data but suppress the noise. The user has the opportunity to inspect the full range of data including the noise at any time. We consider a form of spatially variant filtering of CT volume based upon the differences between the frequency characteristics of the noise and the signal.

There are two major approaches for spatial filtering of CT noise. In the first approach, the projection data is processed (before the reconstruction step)~\cite{king1981noise,huang1980cubic}. In the second approach, the reconstructed data is processed (after the reconstruction step)~\cite{riederer1978noise,henrich1980simple}. We chose the latter approach to have a general solution, which is independent of the diverse requirements of CT scanners~\cite{schaap2008fast}.

\begin{figure*}[!t]
\centering
\includegraphics[width=0.8\textwidth]{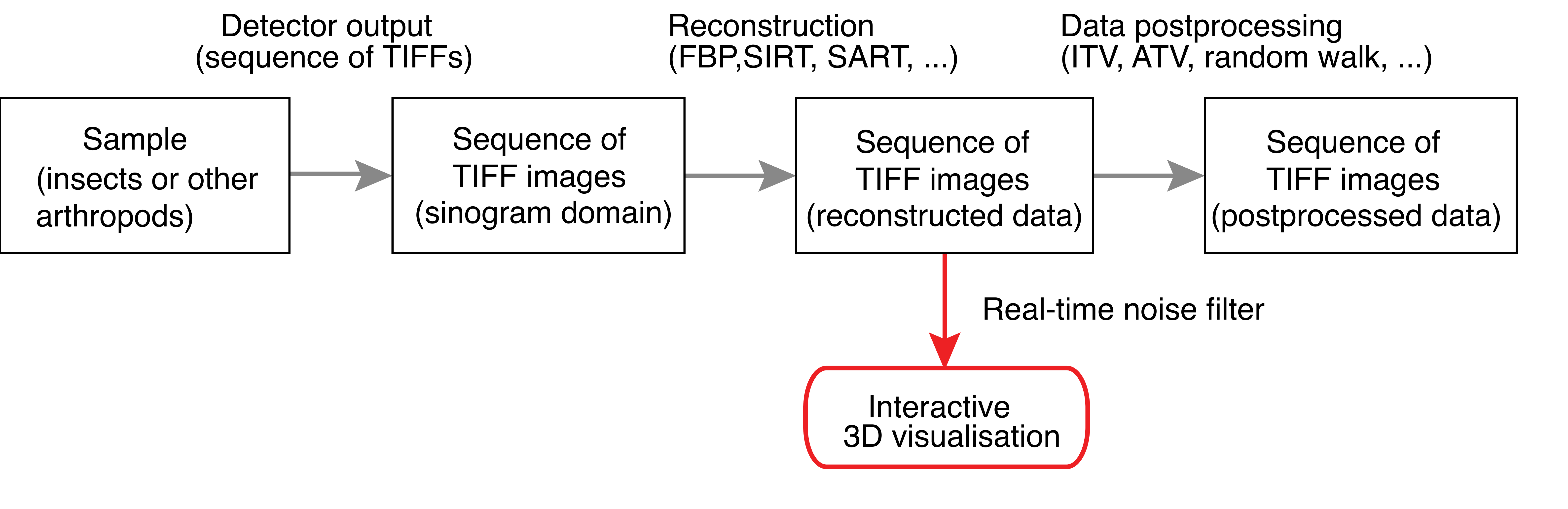}
\caption{Picture of the image processing pipeline for a tomography-oriented scientific workflow.
The red represents the stage, where we enable rapid feedback of the data quality visually.}
\label{fig:workflow}
\end{figure*}

Figure~\ref{fig:workflow} shows the tomographic-oriented scientific workflow in which we highlight where our visualization service fits in the flow (labeled in red). During the data acquisition phase, our service provides rapid feedback on the data quality, allowing users to monitor and adjust experiment setups in real-time. To provide the final interactive 3D surface rendering, we adopted the GPU direct volume rendering approach~\cite{kruger2003acceleration} by terminating the ray casting iteration at the surface intersection point. Since we are using voxel data, we compute the normal vector using the 3D Sobel operator~\cite{sobel19683x3} and apply illumination models on the surface points, i.e., the Phong illumination model~\cite{phong1975illumination}. 

In this paper, we present a local noise filter which takes the diagonally spread neighborhood points to represent the final value.
For each ray iteration, the surface intersection point serves as the central voxel where we perform averaging with its adjacent neighboring voxels---\emph{average cluster}. We repeat the averaging with eight other \emph{average clusters} spread diagonally with a distance of $\sqrt[]{3}$ units from the central voxel. The resulting average is subjected to the data threshold value where any value lower than the threshold is treated as noise. We determine the data threshold value by using the Otsu-threshold method~\cite{otsu1979threshold}, which summarises the data in a binary format. As a result, 
we can suppress the noise in real-time (millisecond range) while preserving the underlying data structure.

\iffalse
\begin{figure*}[!t]
\centering
\includegraphics[width=1.0\textwidth]{placeholder1}
\caption{Motivation: The top images show the final surface 3D visualisation of the tachinid fly \textit{Gymnosoma nudifrons} (Herting, 1966) data set by varying the lower limit of the threshold. The higher the threshold value is, the more details are removed. The lower left image shows the visualisation with threshold value computed by the Otsu-method. The lower middle image shows the fly data set with well-preserved details. However, the 3D visualisation is surrounded by noise. Using additional filtering of the data at threshold 109, the noise is no longer visible but at the expense of removing details---the cuticle of the eyes and wings are not clearly visible.}
\label{fig:motivation}
\end{figure*}
\fi

\section{Related Work}
\label{sec:related}
Most studies of spatial filtering are performed offline with two-dimensional CT images. Since the CT volume is often provided as a stack of 2D cross-sectional images~\cite{hu2001automatic}, various 2D filters are applied to remove the CT noise. We will study these filters and apply them later in the 3D context (Section~\ref{sec:others}). Hence, we will use the term point rather than voxel throughout this section.

Low-pass filtering is the most common filter that replaces each point with the average values of the defined kernel size~\cite{rutherford1976measurement,chew1978effect}. Low-pass filtering removes the high frequencies from the volume data which reduces the noise and improves the detectability of data structure. However, the filter reduces the intrinsic resolution of the data, i.e., smoothes edges and decreases the visibility of small structures. Since the low-pass filter is non-deterministic, an increasing kernel size will lead to the unintended removal of data.

Okada~\cite{okada1985noise} presented an approach based on the assumption that significant differences between neighboring voxels are unlikely to be caused by noise. Okada's approach smoothes only the voxels with a difference to adjacent voxels below a predefined threshold.

Lee et al.~\cite{lee1983digital} presented the sigma filter which assumes a Gaussian distribution of CT noise. They showed the effectiveness of the sigma filter in preserving subtle details and line features as long as the intensity difference between them and their background is higher than the two-sigma intensity range. In other words, the sigma filter only considers grey values that fall within the 2-sigma intensity range and then it performs a straight averaging on the selected values. Similarly, McDonnell~\cite{mcdonnell1981box} presented an extended box-filtering approach that defines the intensity region empirically.

Apart from spatial filtering, we also considered an entropy-based approach which calculates the information entropy of a given kernel~\cite{abutaleb1989automatic,borda2011fundamentals,thum1984measurement}. We use the Shannon-Wiener entropy criterion~\cite{applebaum1996probability} to characterize the noise distribution---average uncertainty of the values---within the kernel. However, the influence of random effects can adversely affect the accuracy of entropy calculation~\cite{likar2000retrospective}.

\section{CT Data and 3D Visualisation}

The tachinid fly \textit{Gymnosoma nudifrons} (Herting, 1966) had been fixed and stored in 70\% EtOH. The unstained sample was scanned using SR\textmu CT at the ANKA synchrotron radiation facility~\cite{schmelzle2017nova,simon2003x}. We used a beam energy of 20 keV and 3000 projections at 250 projections per second. We used a magnification of 1.8x with a resulting field of view of 1.2 mm and an effective pixel size of 6.11 \textmu m. The scanned data (projection data) are later reconstructed and stored on the experiment server.

The 3D interactive visualization system is part of the raycasting framework~\cite{jerome2017wave} that allows users to visualize the reconstructed data from the experiment station. Our visualization approach uses the GPU direct volume rendering approach~\cite{kruger2003acceleration} that performs ray casting on the voxel data. For surface rendering, we terminate the ray iteration at the surface intersection point onto which we then apply illumination models such as the Phong illumination model~\cite{phong1975illumination}. Since we are dealing with voxel data, we approximate the normal vector of the surface point by using the 3D Sobel operator~\cite{sobel19683x3}.

\section{Our Method}
\label{sec:ours}
\begin{figure}[tb]
 \centering % avoid the use of \begin{center}...\end{center} and use \centering instead (more compact)
 \includegraphics[width=\columnwidth]{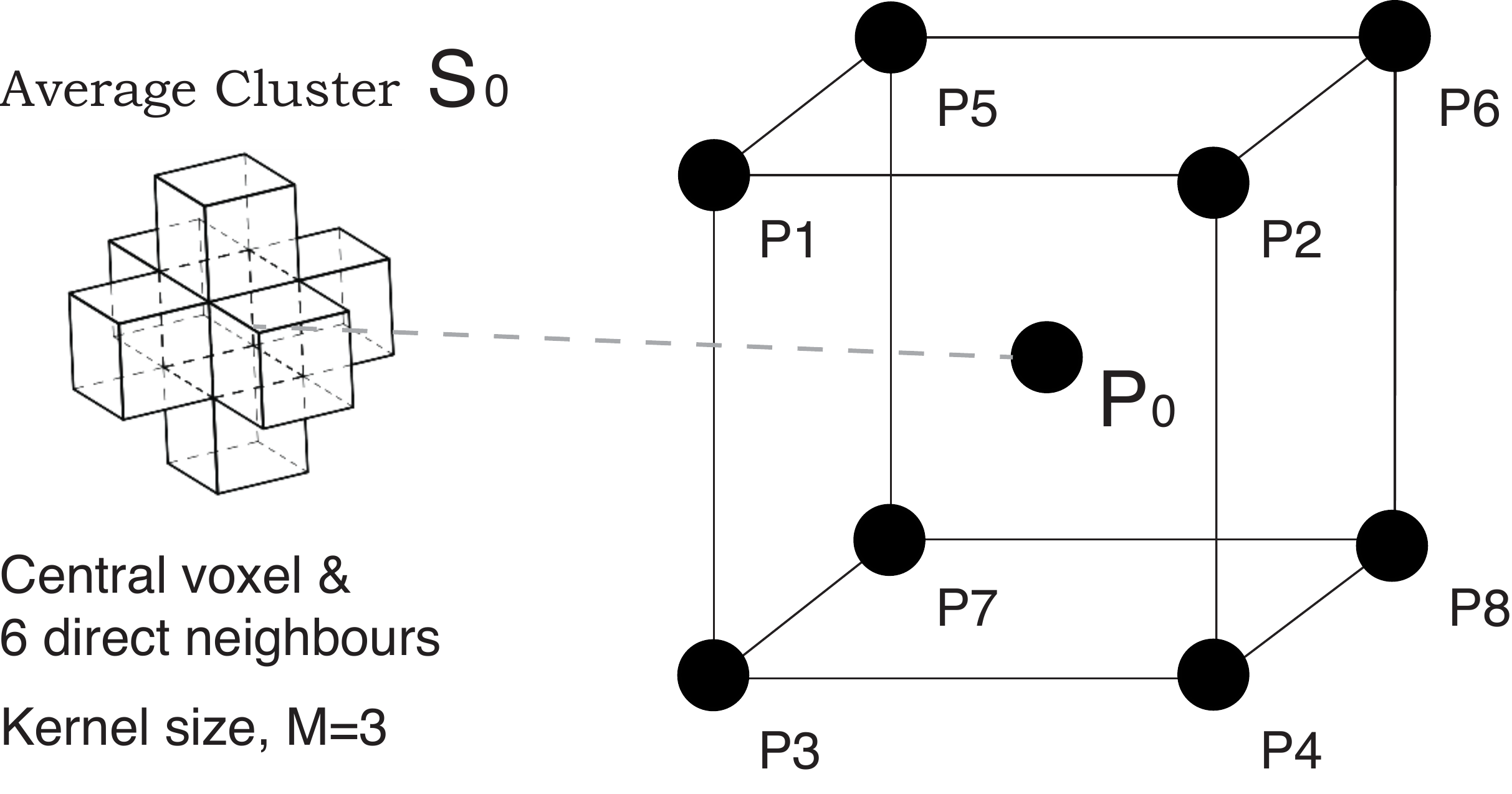}
 \caption{Illustration of the \emph{average cluster} which takes the average of central voxels and its adjacent neighbors. $P$ denotes the voxel, and $S$ represents the \emph{average cluster}. The average of nine \emph{average clusters} ($S_0...S_8$) is used to represent the surface intersection point.}
 \label{fig:star}
\end{figure}

In our approach, we perform spatial filtering in real-time that suppresses the noise at the first surface intersection point within the volume rendering framework. Hence, we consider spatial voxels in the 3D space. 
Most filters (Section~\ref{sec:related}) are applied to 2D cross-sectional images of a volumetric data and remove noise permanently. In contrast, our approach preserves the original data and thus encourages visual data exploration~\cite{keim2001visual}. 

Our method modifies the mean filter because the mean filter tends to provide us with an over-averaged result: using small kernel ($3\times 3\times 3$) cannot suppress spot noise, and larger kernel ($5\times 5\times 5$ or $7\times 7\times 7$) removes details of the data structure. Instead, we average smaller regions ($3\times 3\times 3$) spread around the central kernel to give us a better result. We consider nine small regions, dubbed as \emph{average clusters} ($S_0 \ldots S_8$): 8 regions spread diagonally with a distance of $\sqrt[]{3}$ units from the central kernel ($S_1 \ldots S_8$), and the region at the central kernel itself ($S_0$) (Figure~\ref{fig:star}).

Our goal is to provide the user with the best visual quality without any manual intervention. Along each ray iteration, the local noise filter starts at the surface intersection point and determines whether the position is noise. Firstly, it is vital to decide on the threshold that separates the features of data and the background noise automatically. Secondly, we aim to provide the best visual quality possible by suppressing the noise artifacts and at the same time preserving the features and structures of the data. 

%Note that we do not remove any data during the filtering, but instead not showing the noise during the visualization session. 

To determine the threshold of the data, we applied the Otsu thresholding method which scans through the grey levels to find the threshold that minimizes the intra-class variance---weighted sum of variances of the two classes~\cite{otsu1979threshold}

\begin{figure*}[tb]
 \centering % avoid the use of \begin{center}...\end{center} and use \centering instead (more compact)
 \includegraphics[width=1.0\textwidth]{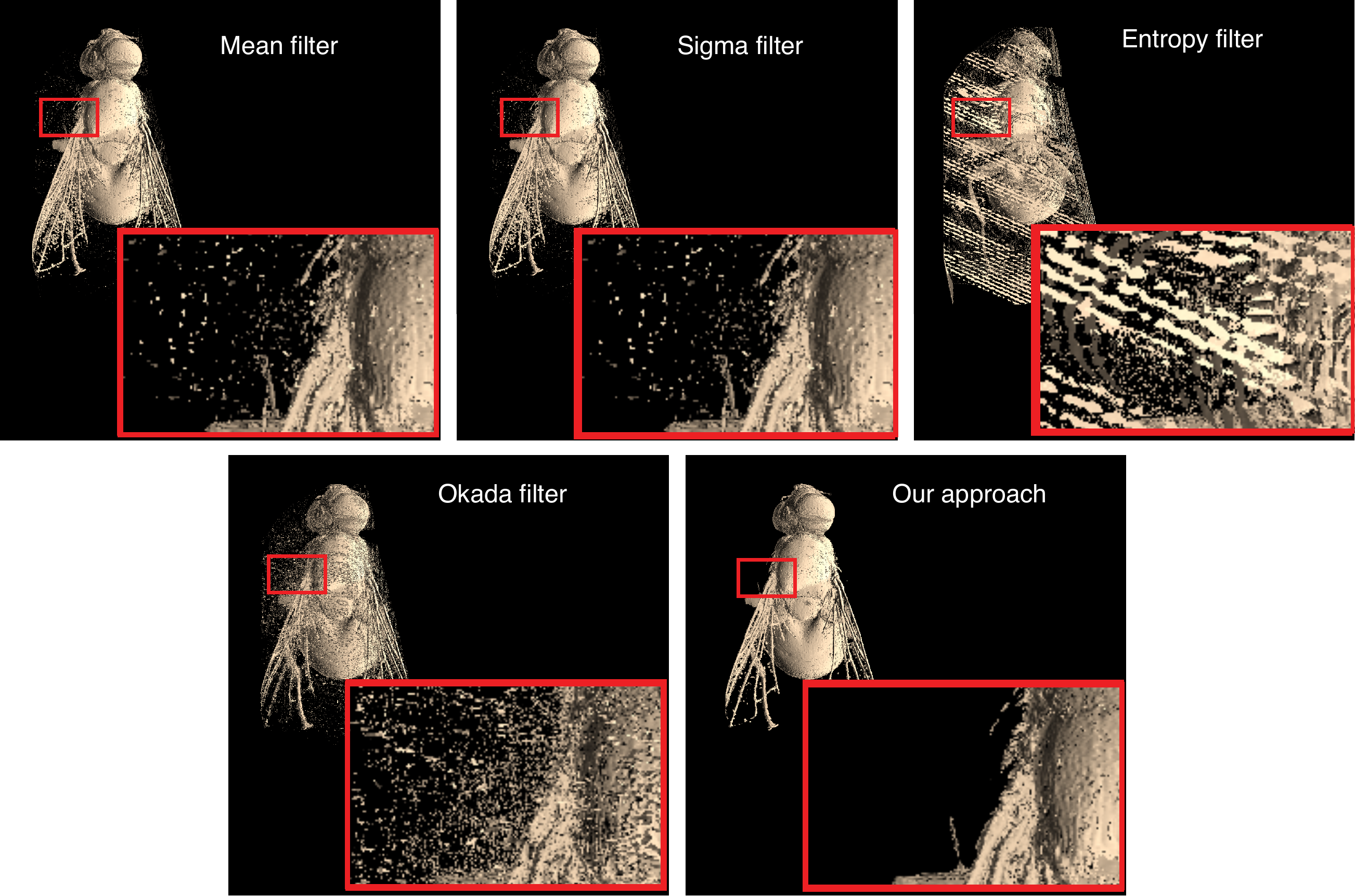}
 \caption{A visual comparison of the tachinid fly Gymnosoma nudifrons (Herting, 1966) data set
between the mean filter (top left), the sigma filter
(top middle), the entropy filter (top right), the Okada filter (bottom left) and our approach (bottom
right). The entropy filter fails to suppress the noise completely. Also, there is still spot noise in
the mean, sigma and Okada filters which our approach can suppress.}
 \label{fig:visual}
\end{figure*}

\begin{equation}
\sigma^{2}_{\omega}(T) = \omega_{0}(T)\sigma^{2}_{0}(T) + \omega_{1}(T)\sigma^{2}_{1}(T)
\end{equation}

\noindent where $\omega _{0}$ and $\omega _{1}$ are the weights which represent the probabilities of the two classes separated by the threshold $T$. $\sigma _{0}^{2}$ and $\sigma _{1}^{2}$ are the variances of the two classes. We use the Otsu threshold as the lower limit of the intensity range.

After filtering the data with the Otsu-threshold $T$, we detect a wide-array of surface intersection points. Each intersection surface point serves as the central voxel where we determine whether the corresponding voxel is noise or not. Here, we compute the average value based on the central voxel together with its adjacent neighbouring voxels within the kernel size, $M$:

\begin{equation}
\begin{split}
S_0 = \frac{1}{3\cdot M}\Biggl( & \sum_{i=0}^{M-1} P(s_x+\Delta i,s_y,s_z)\ + \\  
                         & \sum_{i=0}^{M-1} P(s_x,s_y+\Delta i,s_z)\ + \\
                         & \sum_{i=0}^{M-1} P(s_x,s_y,s_z+\Delta i) \Biggr),
\end{split}
\end{equation}

with 

\begin{equation}
\Delta i = i- \frac{M-1}{2}.
\end{equation}

%\begin{equation}
%S_0 = \frac{1}{7}\left(P_0 + \sum_{i=0}^{M/3}\sum_{j=0}^{M/3}\sum_{k=0}^{M/3} P(s_x+\Delta i,s_y+\Delta j,s_z+\Delta k)\right),
%\end{equation}

\noindent The $P_0$ is the grey value of the central voxel with a spatial coordinate of $(S_x,S_y,S_z)$. The $\Delta i$ represents the offset value from the central voxel. Let $S_0$ be the \emph{average cluster} at the central voxel, we further take eight additional \emph{average clusters} ($S_1 ... S_8$) spread diagonally around the central voxel with a distance of $\sqrt{3}$ units. We replace the corresponding voxel value with the resulting average value from the nine clusters.

\section{3D Spatial Noise Filters}
\label{sec:others}

For completeness, we describe how we adapted other spatial noise filters (Section~\ref{sec:related}) in our 3D visualization framework. In particular, we consider low-pass filtering (mean filter), Okada filter, sigma filter and the entropy filter. Similar to our method, the filters start at the surface intersection point.

\paragraph{Mean Filter}

We calculate the average of voxel values that lies within the kernel size, $M$:

\begin{equation}
S = \frac{1}{M^3}\left(\sum_{i=0}^{M-1}\sum_{j=0}^{M-1}\sum_{k=0}^{M-1} P(s_x+\Delta i,s_y+\Delta j,s_z+\Delta k)\right),
\label{eq:s0}
\end{equation}

\noindent where $P$ is the grey value of the voxel with the spatial coordinate of $(S_x,S_y,S_z)$. The $\Delta i$ represents the offset value from the central voxel which is defined as $(i*1-1)$.

% kernel size, n = 3

\paragraph{Sigma Filter}

Within the predefined kernel size $M$, the sigma filter only considers $n$ voxel values that fall within the intensity range specified by the global standard deviation. Specifically, the intensity range must be within the $2\sigma$ region:

\begin{equation}
    S = \frac{1}{n}\sum_{i=0}^{n-1} P_i,\ \ \ \ \forall P_i \in {[{-2\sigma},2\sigma]}.
\end{equation}

%\begin{equation}
%    S = \frac{1}{n}\sum_{i=0}^{n-1} P_i,\ \ \ \ with\ \ P_i>-2\sigma \wedge %P_i<2\sigma.
%\end{equation}

% kernel size, n
% sigma range (LL, UL)

\paragraph{Okada Filter}

The Okada filter studies the difference between the central voxel value, $P_0$, with its neighboring voxel values, $P_{i}$. We consider the neighboring voxels only when the difference value is lower than the predefined threshold $T_d$:

\begin{equation}
    S=
    \begin{cases}
      \frac{1}{n}\sum_{i=0}^{n-1} P_i, & \text{if}\ \lvert P_0 - P_{i}\rvert < T_d \\
      0, & \text{otherwise},
    \end{cases}
\end{equation}

\noindent where $n$ is the total number of voxels that satisfy the condition $\lvert P_0 - P_{i}\rvert < T_d$.
  
% However, the intensity difference between the voxels for noise and features overlaps, hence we cannot differentiate the noise only by judging the several pairs of voxels.

% D

\paragraph{Entropy Filter}

We calculate the first order entropy $H$ from $m$ voxels within the kernel size $M$. Then, we set an entropy threshold $T_e$ to determine the noise range. Before filtering, we precompute the normalized frequency distribution of the data to assign a probability $p_i$ to each grey level. If the entropy criterion~\cite{thum1984measurement} exceeds our predefined threshold, then we choose to display the central voxel value, $P_0$.

\begin{equation}
    S=
    \begin{cases}
      P_0, & \text{if}\ - \sum_{i=0}^{m-1} p_i \log _{2}\left(p_i\right) >T_e \\
      0, & \text{otherwise}.
    \end{cases}
\label{eq:entropy_filter}
\end{equation}

% Te =
% kernel size, n

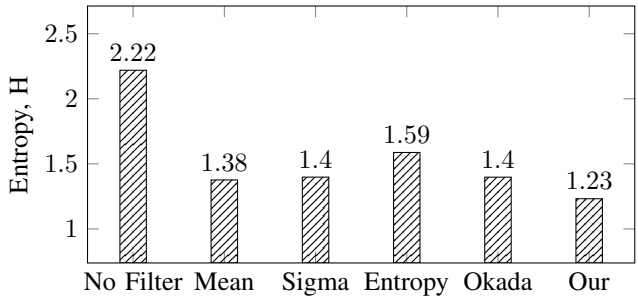
\begin{figure}
\begin{tikzpicture}
    \begin{axis}[
            symbolic x coords={No Filter,Mean,Sigma,Entropy,Okada,Our},
            xtick=data,
            ylabel={Entropy, H},
            ylabel style={yshift=-1.0em},
            height=5cm,
            width=\columnwidth,
            nodes near coords,
            enlarge y limits=0.5
          ]
            \addplot[ybar,pattern=north east lines] coordinates {
                (No Filter,   2.2195)
                (Mean,  1.3766)
                (Sigma,   1.3982)
                (Entropy,   1.5876)
                (Okada,   1.3978)
                (Our,   1.2322)
            };
    \end{axis}
\end{tikzpicture}
\caption{Visual comparison: Entropy value of the final 3D images produced without any filter, with the mean filter, the sigma filter, the entropy filter, the Okada filter and our approach. The lower the entropy value, the better.}
 \label{fig:barfilter}
\end{figure}

\section{Results and Discussion}

The visualization framework is written in C++ OpenGL. We included the filters described in Section~\ref{sec:ours} and ~\ref{sec:others} where we evaluated the visual quality and performance of each filter respectively. Given the various tuning parameters for each filter, we select the settings that suppress the most noise whilst preserving fine details of the sample data. We will use a kernel size of 3 ($M=3$) throughout our evaluation.

\subsection{Visual Quality}

%To evaluate the resulting quality of the filters, we compare the results to a manually segmented data set that serves as our ground truth. The reference data is a surface mesh model that is generated in Amira\textsuperscript{\textregistered} 5.6.0 (FEI, Munich, Germany). 
%based on a grey value threshold of 104. 

Figure~\ref{fig:visual} shows the visual quality of the tachinid fly data set. For this particular data set, we calculated the Otsu-threshold $T$ as 101, and we will apply this threshold setting across the filters.

Among the filters applied, the entropy filter performs the worst because the noise and the useful data regions are overlapping. Since the data is preprocessed by Amira\textsuperscript{\textregistered} 5.6.0 (FEI, Munich, Germany), the manually sliced dataset produces interpolation artifacts at the boundary layers. On the other hand, the mean filter, the sigma filter, and the Okada filter can suppress the noise considerably and outline the tachinid fly details. However, stubborn spot noise is surrounding the object.

%Interestingly, the fine structures of the wings are not readily identifiable even by the domain experts.

In our approach, we included not only the average information at the central voxel but also eight uniformly spread \emph{average clusters} for better noise-to-information analysis. As a result, we suppress most of the noise including spot noise. We further evaluated the final visual quality by calculating the entropy value on the resulting 3D images (Figure~\ref{fig:barfilter}). The lower the entropy value, the better is the visual quality. Our approach has the lowest entropy value which is a clear indication regarding the effectiveness of our filter in suppressing even spot noise.

\subsection{Performance}
\begin{figure}
\begin{tikzpicture}
    \begin{axis}[
            symbolic x coords={Mean,Sigma,Entropy,Okada,Our},
            xtick=data,
            ylabel={Time per frame, ms},
            ylabel style={yshift=-1.0em},
            height=5cm,
            width=\columnwidth,
            nodes near coords,
            enlarge y limits=0.5,
          ]
            \addplot[ybar,pattern=north east lines] coordinates {
                (Mean,  43)
                (Sigma,   48)
                (Entropy,   50)
                (Okada,   44)
                (Our,   53)
            };
    \end{axis}
\end{tikzpicture}
\caption{Performance: Time measurement per frame of the mean filter, the sigma filter, the entropy filter, the Okada filter and our approach. Less time indicates a better performance.}
 \label{fig:bartime}
\end{figure}
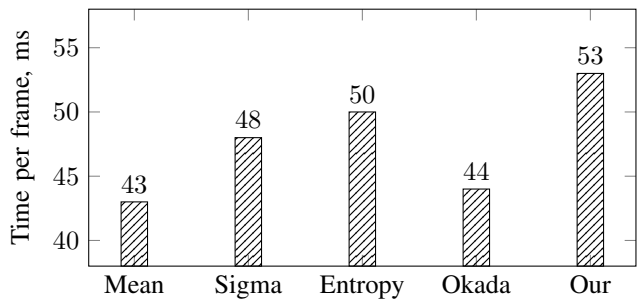

We measured the performance of the filters (Figure~\ref{fig:bartime}) using an Intel\textsuperscript{\textregistered} Core\textsuperscript{\texttrademark} i5-4670 CPU (4 x 3.40GHz) with an NVIDIA Tesla C2070. The mean filter is the fastest due to its simplicity in taking all voxel values within the kernel and performing a straight averaging. The Okada filter involves constant checking of the difference value between the neighboring voxel and the central voxel which leads to a slower time. Also, the Okada filter only takes the six direct adjacent voxels into consideration. On the other hand, the sigma filter considers all the voxels within the defined kernel size, where its grey values fall within the $2\sigma$ range. The entropy filter performs an entropy computation (Equation~\ref{eq:entropy_filter}) that increases the overall complexity and hence the time needed for computation.

The time per frame for our approach is the highest among the filters because of the extra \emph{average cluster} fetching around the central voxel. The idea is to cover a broader spatial region to identify the characteristic of the current central voxel accurately. Our approach takes longer in comparison to the other spatial filters. The visual quality, however, is significantly improved over the other filters while taking only \SI{10}{\milli\second} longer than the (fastest) mean filter.

\section{Conclusions and Future Work}

We presented a local noise filter which takes the diagonally spread neighborhood points to represent the final value. Our filter is a combination of Otsu-threshold and the extended mean filter methods to provide an automatic 3D visual rendering of CT data and does not require user intervention. The resulting 3D image suppresses even spot noise which was not possible with other filters. Although our approach is similar to the mean filter, we take more \emph{average clusters} to improve our final average value. The approach keeps the processing time within the millisecond range (\SI{53}{\milli\second} to render a frame) making it suitable for an interactive visualization system.

For future work, we want to test our filter with a broader set of CT data. We would like to study the effectiveness of our filter to maintain the fine structure of small animals (for example arthropods). For now, we evaluated our filter with eight \emph{average clusters} uniformly spread around the central voxel with a unit distance of $\sqrt{3}$. It would be interesting to analyze several variations and study the effects of each combination.

\section*{Acknowledgment}

Data and/or analytical tools used in this study were provided by the projects ASTOR and NOVA (Michael Heethoff, TU Darmstadt; Vincent Heuveline, Heidelberg University; J\"{u}rgen Becker, Karlsruhe Institute of Technology), funded by the German Federal Ministry of Education and Research (BMBF; 05K2013, 05K2016). We especially thank the following co-workers: Felix Beckmann, J\"{o}rg Hammel, Andreas Kopmann, Philipp L\"{o}sel, Wolfgang Mexner, Tomy dos Santos Rolo, Nicholas Tan Jerome, Matthias Vogelgesang, Tom\'{a}\u{s} Farag\'{o}, Sebastian Schmelzle, Thomas van de Kamp.

\ifCLASSOPTIONcaptionsoff
  \newpage
\fi



\bibliographystyle{IEEEtran}
%\bibliographystyle{abbrv-doi}
\bibliography{IEEEabrv,reference}
\end{document}